\documentclass[conference]{IEEEtran}
\IEEEoverridecommandlockouts
\usepackage{cite}
\usepackage{amsmath,amssymb,amsfonts}
\usepackage{graphicx}
\usepackage{textcomp}
\usepackage{xcolor}
\usepackage{titlesec}
\def\BibTeX{{\rm B\kern-.05em{\sc i\kern-.025em b}\kern-.08em
    T\kern-.1667em\lower.7ex\hbox{E}\kern-.125emX}}
\usepackage{url}
\usepackage{algorithm}
\usepackage{diagbox}
\usepackage{multirow}
\usepackage[acronym]{glossaries}
\usepackage{algpseudocode}

\glsdisablehyper
\newacronym{AI}{AI}{Artificial Intelligence}
\newacronym{IoT}{IoT}{Internet of Things}
\newacronym{ANN}{ANN}{Artificial Neural Network}
\newacronym{SNN}{SNN}{Spiking Neural Network}
\newacronym{DNN}{DNN}{Deep Neural Network}
\newacronym{RNN}{RNN}{Recurrent Neural Network}
\newacronym{CNN}{CNN}{Convolutional Neural Network}
\newacronym{MCU}{MCU}{MicroController Unit}
\newacronym{CL}{CL}{Continual Learning}
\newacronym{SHD}{SHD}{Spiking Heidelberg Dataset}
\newacronym{LR}{LR}{Latent Replay}
\newacronym{LIF}{LIF}{Leaky Integrate and Fire}
\newacronym{SG}{SG}{Surrogate Gradient}
\newacronym{BPTT}{BPTT}{Back-Propagation Through Time}

\usepackage{titlesec}
\titlespacing*{\section}{0pt}{0.3\baselineskip}{0.1\baselineskip}
\titlespacing*{\subsection}{0pt}{0.3\baselineskip}{0.1\baselineskip}

\usepackage[belowskip=-14pt,aboveskip=0pt]{caption}

\begin{document}

\title{Compressed Latent Replays for Lightweight Continual Learning on Spiking Neural Networks}

\author{
    Alberto Dequino$^{1,2}$,
    Alessio Carpegna$^2$,
    Davide Nadalini$^{1,2}$,
    Alessandro Savino$^2$,
    Luca Benini$^{1, 3}$, \\
    Stefano Di Carlo$^2$,
    Francesco Conti$^{1}$ 
    
    ($^1$Universit\'a di Bologna,
    $^2$Politecnico di Torino,
    $^3$ETH Zurich )

    \\[-5.0ex]
}

%\and
%\and
%\and
%\and
%\and

\maketitle

\begin{abstract}
    Rehearsal-based Continual Learning (CL) has been intensely investigated in Deep Neural Networks (DNNs). However, its application in Spiking Neural Networks (SNNs) has not been explored in depth. In this paper we introduce the first memory-efficient implementation of Latent Replay (LR)-based CL for SNNs, designed to seamlessly integrate with resource-constrained devices. LRs combine new samples with latent representations of previously learned data, to mitigate forgetting.  
    Experiments on the Heidelberg SHD dataset with Sample and Class-Incremental tasks reach a Top-1 accuracy of 92.5\% and 92\%, respectively, without forgetting the previously learned information. Furthermore, we minimize the LRs’ requirements by applying a time-domain compression, reducing by two orders of magnitude their memory requirement, with respect to a na\"ive rehearsal setup, with a maximum accuracy drop of 4\%. On a Multi-Class-Incremental task, our SNN learns 10 new classes from an initial set of 10, reaching a Top-1 accuracy of 78.4\% on the full test set
    \footnote{To encourage research in this field, we release the code related to our experiments as open-source: \url{https://github.com/Dequino/Spiking-Compressed-Continual-Learning}}.
\end{abstract}


% ------------ INTRODUCTION ------------
\section{Introduction}

\color{black}
%In recent years, hardware and software advancements in edge \gls{IoT} devices have enabled the integration of \gls{AI} on low-power sensor nodes. 
%The current landscape of edge AI is dominated by \glspl{ANN} driven by high-end server models, where Transformers excel in Natural Language Processing, and \glspl{CNN} are efficient in image and video analysis tasks \cite{10098596}. 
%
%\color{blue}The recent advancements in hardware and software for edge \gls{IoT} devices have enabled the integration of \gls{AI} on low-power sensor nodes, inspired by high performances of server-driven \glspl{ANN} models, like Transformers for Natural Language Processing and \glspl{CNN} for image and video analysis \cite{10098596}.
%
The current landscape of edge AI is dominated by \glspl{ANN} driven by high-end server models, like Transformers and \glspl{CNN} \cite{10098596}. Concurrently, the recent advancements in hardware and software for edge \gls{IoT} devices have enabled the integration of \gls{AI} on low-power sensor nodes.
% The current landscape of edge AI is dominated by \glspl{ANN} driven by high-end server models, where Transformers excel in Natural Language Processing, and \glspl{CNN} are efficient in image and video analysis tasks \cite{10098596}. 
%
However, deploying complex \glspl{ANN} for edge inference involves imposing constraints like quantization and pruning \cite{liang2021pruning} to accommodate small \gls{IoT} devices, like ultra-low-power microcontrollers. 
\color{black}
Moreover, edge \gls{AI} models are susceptible to errors, once deployed, due to shifts in data distribution between training and operational environments \cite{amodei2016concrete}. Also, an increasing number of applications require adapting \gls{AI} algorithms to individual users while maintaining privacy and minimizing %reliance on 
internet connectivity.

\gls{CL} — i.e., the ability to continually learn from evolving environments, without forgetting previously acquired knowledge — emerges as a novel paradigm to address these challenges. %Two primary approaches have been explored extensively in \glspl{ANN}.
%\cite{9349197}
\textit{Rehearsal-based} methods \cite{Rebuffi_2017_CVPR,9341460}, the most accurate \gls{CL} solutions, mitigate forgetting by continually training the learner on a mix of new data and a stored set of samples from previously learned tasks, albeit at the expense of additional on-device storage. \textit{Rehearsal-free} methods \cite{ehret2021continual,lomonaco2020rehearsal} rely on tailored modifications to network architecture or learning strategy to ensure model resilience to forgetting, without saving samples on-device, but with a potential trade-off in accuracy.
\gls{CL} at the edge, especially Rehearsal-based methods, can be resource-intensive for various \gls{ANN} models, as \glspl{CNN}, %(e.g., \glspl{CNN}, or \glspl{RNN})
demanding substantial on-device storage for complex learning data. %Additionally, many models often require the execution of the backpropagation algorithm at the edge.

In this context, \glspl{SNN} emerge as a promising energy-efficient paradigm, exhibiting high accuracy and efficiency in processing time series \cite{10.1007/978-3-030-58607-2_23}. \glspl{SNN} closely emulate biological neurons' behavior, communicating with spikes that can be efficiently stored as 1-bit data in digital architectures. This simplified data encoding creates new opportunities for developing \gls{CL} solutions. While online learning has been explored in both hardware and software \glspl{SNN} implementations \cite{9731734}, \gls{CL} strategies in \glspl{SNN} are only partially investigated in Rehearsal-free methods \cite{10.3389/fncom.2022.1037976,9586281,han2023adaptive}. 
%To the best of our knowledge, Proietti, et al. \cite{proietti2023memory} are the only ones who addressed Rehearsal-based \gls{CL} in \glspl{SNN}. However, their work shows limited accuracy and does not optimize memory storage, which is vital for edge devices.
To the best of our knowledge, only Proietti, et al. \cite{proietti2023memory} addressed Rehearsal-based \gls{CL} in \glspl{SNN}. However, their work shows limited accuracy and does not optimize memory storage, which is vital for edge devices.
\color{black}
This work makes the following contributions:
\begin{itemize}
    \item A cutting-edge, memory efficient implementation of Rehearsal-based \gls{CL} for \glspl{SNN}, designed to seamlessly integrate with resource-constrained devices. We enable \gls{CL} on \glspl{SNN} by exploiting a Rehearsal-based algorithm — i.e., \textit{\gls{LR}} — proving to achieve State-of-the-Art classification accuracy on \glspl{CNN} \cite{9341460}.
    \item A novel approach to reduce the rehearsal memory, based on the robustness of information encoding in \glspl{SNN} to precision reduction, which allows us to apply a lossy compression on the time axis.
    \item We validate our method targeting a keyword spotting application, using Recurrent \gls{SNN}, on two common \gls{CL} setups: \textit{Sample-Incremental} and \textit{Class-Incremental} \gls{CL}. 
    \item Finally, to highlight the proposed approach's robustness, we test it in an extensive Multi-Class-Incremental \gls{CL} routine, learning 10 new classes from an initial set of 10 pre-learned ones. %achieving a Top-1 accuracy of 78.4\% on the full test set.
\end{itemize}

% VAlidation on Sample and Class Incremental
In the \textit{Sample-Incremental} setup, we achieve a Top-1 accuracy of \textcolor{black}{92.46\%} on the \gls{SHD} test set, requiring 6.4 MB for the \glspl{LR}. This occurs while incorporating a new scenario, for which the accuracy is improved by 23.64\%, without forgetting the previously learned ones. In the Class-Incremental setup, a Top-1 accuracy of \textcolor{black}{92\%} was attained, requiring 3.2 MB, when learning a new class with an accuracy of 92.50\%, accompanied by a loss of up to 3.5\% on the old classes. 
When jointly applying compression and selecting the most performing \gls{LR} index, we reduce by 140$\times$ the memory requirements for the rehearsal data, while losing only up to 4\% accuracy, with respect to the na\"ive approach. 
Moreover, in the Multi-Class-Incremental setup, we achieve 78.4\% accuracy with compressed rehearsal data, while learning the set of 10 new keywords. These results pave the way to a new low-power and high-accuracy approach for \gls{CL} on edge.
\color{black}

\section{Related Work}

\gls{CL} on \glspl{ANN} involves two main approaches: Rehearsal-based and Rehearsal-free methods.
In Rehearsal-based methods, mitigating catastrophic forgetting involves training the learner on a mix of newly acquired data and samples from previously learned tasks. To increase learned classes, iCaRL \cite{Rebuffi_2017_CVPR} utilizes a set of representative old class examples as rehearsal data, chosen to maintain a balanced set of classes. To improve the efficiency of Rehearsal-based methods, Pellegrini et al. \cite{9341460} propose storing rehearsal data as \glsentrylongpl{LR}, i.e., activations produced as the output of one of the hidden layers of the learner, specifically a feed-forward \gls{CNN}. Only the last layers are retrained, while the earliest backbone's weights are frozen. Results, compared to iCaRL, demonstrate three orders of magnitude faster execution, with an almost 70\% improvement in Top-1 accuracy on a Class-Incremental setup using the CORe50 dataset.
In Rehearsal-free methods, addressing forgetting involves modifying the learner's architecture or customizing the training procedure \cite{ehret2021continual,lomonaco2020rehearsal}. However, their accuracy lags behind Rehearsal-based approaches.
We adopt the Rehearsal-based approach using \glspl{LR}, given its demonstrated effectiveness on \glspl{CNN}.

Continual Learning in \glspl{SNN} has primarily focused on Rehearsal-free methods. Skatchkovsky et al. \cite{10.3389/fncom.2022.1037976} propose a Bayesian continual learning framework, providing well-calibrated uncertainty quantification estimates for new data. In contrast, the SpikeDyn framework \cite{9586281} introduces unsupervised continual learning on a model search algorithm, supporting adaptive hyperparameters. However, this approach performs poorly, achieving 90\% accuracy on average for a new class in the MNIST dataset while only maintaining 55\% accuracy on the old classes. Drawing inspiration from human brain neurons, Han et al. \cite{han2023adaptive} propose Self-Organized Regulation networks, capable of learning new tasks by self-organizing the neural connections of a fixed architecture. Additionally, they simulate irreversible damage to \glspl{SNN} structures by pruning the models. Results on the CIFAR100 and Mini-ImageNet datasets, using DNN-inspired Convolutional and Recurrent networks, demonstrate an average accuracy of around 80\% and 60\%, respectively, for all learned classes.
To the best of our knowledge, only the work of Proietti, et al., \cite{proietti2023memory} targets Rehearsal-based \gls{CL} on \glspl{SNN}. In their approach, a Convolutional SNN model is trained to learn in a Class-Incremental and Task-free manner on the MNIST dataset, learning multiple binary classification tasks in sequence. 
\color{black}
However, their exploration doesn't involve any memory optimization technique, storing raw data for the rehearsal phase. Additionally, they report a top-1 accuracy of 51\% after learning sequentially the 10 target classes, while requiring 16MB of rehearsal memory. 

% However, they demonstrate a classification accuracy of more than 99\% on the latest learned tasks, while older ones are progressively forgotten to a minimum of 49.9\%. 

To improve accuracy and optimize memory efficiency, we explore the reharsal-based domain, applying efficient time-compressed \glspl{LR} to \glspl{SNN}. 
\color{black}
%On the contrary, \textcolor{red}{noi riusciamo, nonostante il diverso dataset, a mantenere un'accuratezza simile, con però un'ottimizazione anche della memoria.}

%As a comparison with the previous work, our fully spiking \gls{SNN} targeting a keyword spotting task on the \gls{SHD} dataset reaches a top 1 accuracy of 78.4\% on a sequential 10-classes incremental learning set-up, with a memory requirement of 12MB.

% ------------ BACKGROUND ------------

\section{Background}
\label{sec:snn_background}

\begin{figure}[t]
		\centering
		\includegraphics[width=0.9\columnwidth]{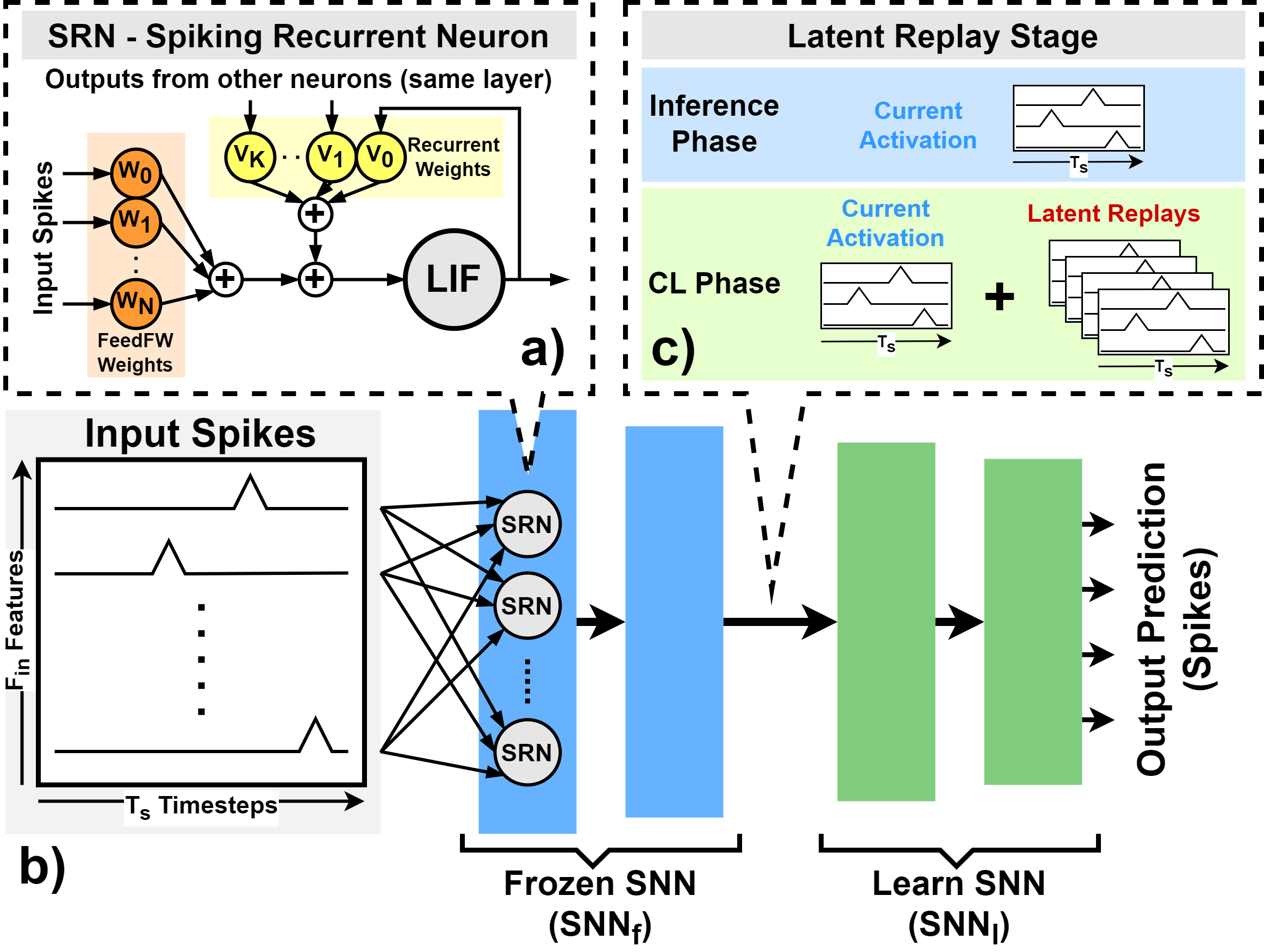}
		\caption{Visual depiction of a generic SNN with Recurrent Neurons and Latent Replays. In a), our Recurrent Neuron model; in b), the structure of a generic Fully-Connected model for Continual Learning; in c), an example of the data fed to the Latent Replay stage.}
		\label{fig:latent_replays}
	\end{figure}

\glspl{SNN} are a category of neural networks inspired by the behavior of biological brain segments. Unlike other \glspl{ANN} models, \glspl{SNN} convey information through sequences of spikes, with their representation adapting to the execution domain.
%In analog domains, spikes may manifest as physical current spikes. Conversely, 
%
In the digital domain, considered in this study, spikes are encoded as binary values (one for a spike, zero otherwise) associated with discrete timesteps. The information conveyed by spikes is encoded in their sequence, such as their instantaneous rate, arrival time, or more complex patterns.
Following Eshraghian, et al., \cite{eshraghian2023} and aiming at the classification of signals with dense time-domain information (e.g., sound samples), we use a Fully Connected Recurrent \gls{SNN} architecture with $L$ layers, incorporating intra-layer side connections akin to those found in \cite{9311226} and \cite{Yin2021}. The chosen model employs a Synaptic Conductance-based Second Order \gls{LIF} neuron model \cite{brette2005}.
Fig. \ref{fig:latent_replays}-a) illustrates the structure of the recurrent neurons, showcasing a collection of feed-forward weights denoted as $W$ linked to each input and a set of recurrent weights labeled as $V$,  each one connected to the output of the neurons within the same layer, including itself.
Optimally training \gls{SNN} models poses a challenge, due to the non-differentiability of the spike function. Similar to \glspl{RNN}, state-of-the-art back-propagation techniques can be applied to \glspl{SNN} using \glspl{SG} \cite{neftci2019}, implemented through \gls{BPTT} \cite{werbos1990}. In this study, \glspl{SNN} undergo training via \gls{BPTT}, utilizing a fast-sigmoid \gls{SG} to approximate the step function characteristic of Recurrent Neurons.

In this work, we target \gls{CL}, a machine learning paradigm that enables models to continuously acquire knowledge from a data stream without erasing prior learning. Let us consider a dataset $D$ on which the model is pre-trained on $K_{class} < N_{class}$ classes and/or $K_{scene} < N_{scene}$ potential scenarios. A scenario here represents a specific data subset containing all $N_{class}$ classes, exemplifying a distinct data variation (e.g., a single speaker in a keyword-spotting dataset). Let $T$ be a test set comprising samples from $N_{class}$ classes and $N_{scene}$ scenarios. In the CL paradigm, the pre-trained model continually incorporates and learns new labeled samples $\not\in D$, enhancing its predictive accuracy on $T$.
CL algorithms require specific techniques to prevent models from forgetting previously learned data, a challenge known as \textit{Catastrophic Forgetting} \cite{Kemker_McClure_Abitino_Hayes_Kanan_2018}, which has recently been demonstrated to exist also in SNNs \cite{hammouamri:hal-03887417, Golden688622}.

% ------------ INCREMENTAL LEARNING ------------
\section{Latent Replay-based Continual Learning in \glspl{SNN}}

In this paper, we draw inspiration from the work of Pellegrini et al. \cite{9341460} on \gls{LR}, initially designed for \glspl{CNN}. Several crucial steps must be addressed to adapt this paradigm for \glspl{SNN}. First, rehearsal data need to encapsulate the temporal evolution of a layer of spiking neurons, in contrast to the static input images in the original \gls{LR} framework. Second, this temporal evolution must fit into the learning algorithm, i.e., \gls{BPTT}, while mixing with newly acquired data. Furthermore, our attention is directed towards two \gls{CL} scenarios: a Sample-Incremental setup, where a learner is trained to classify unseen scenarios, and a Class-Incremental setup, where the learner is tasked with identifying an increasing number of classes.

\subsection{Latent Replays in \glspl{SNN}}

Algorithm \ref{alg:training} outlines the proposed Latent Replay-based training for an \gls{SNN} with $L$ layers. The neural network undergoes pre-training over $E_{pre}$ epochs using \gls{BPTT} on an initial training set $TR_{pre}$ with an $\eta$ learning rate (lines 4-8). As illustrated in Fig. \ref{fig:latent_replays}-b), the network is then split into two sections: ($\mathrm{SNN_{f}}$), comprising the first $K$ layers denoted as \emph{frozen layers}, and ($\mathrm{SNN_{l}}$), encompassing layers from the $L-K+1$-th to the $L$-th, marked as \emph{learning layers} (line 11).
The latent replays (Fig. \ref{fig:latent_replays}-c), denoted as $LR$, constitute a collection of the output activations of the $K^{th}$ layer when exposed to a subset $TR_{replay}$ of the pre-training set. This collection must be stored for later use during \gls{CL} training (line 12), necessitating onboard storage. Since input data are trains of spikes  (i.e., binary values) distributed over $T_s$ time-steps, the stored \glspl{LR} are in turn sequences of $T_s$ single-bit activations.

When training the network on new data, belonging to a new scenario or class, only the \emph{learning layers}, are trained: the \emph{frozen layers} simply propagate the input spikes sequences in the forward direction, processing them through the function learned in the pre-training phase (line 16). The \emph{learning layers} are then trained for $E_{cl}$ epochs, using the output activations of the $K^{th}$ layer, blended with the stored \glspl{LR} to keep memory of the learned data (line 17).

\begin{algorithm} [t]
  \caption{\gls{SNN} Latent Replay training}\label{alg:train_snn}
  \begin{algorithmic}[1]
    \State \textbf{Input:} $TR_{pre}$, $TR_{cl}$.
    \State \textbf{Parameters:} $E_{pre}$, $E_{cl}$, K, $\eta$.
    \\
    \State{\textbf{Initial Training phase:}}
    
    \State Initialize SNN weights $W$, $V$ randomly
    \For{$e \gets 1$ to $E_{pre}$}
      \State ($W$, $V$) $\gets$ BPTT.train(SNN,$TR_{pre}$,($W$, $V$), $\eta$)
    \EndFor

    \\
    \State{\textbf{Prepare network for CL:}}
    \State ($\mathrm{SNN_{f}},\mathrm{SNN_{l}}$) = split (SNN, $K$)
    \State $LR$ = Inference($TR_{replay} \subseteq TR_{pre} $, $\mathrm{SNN_{f}}$)  

    \\
    \State{\textbf{Train network on new data:}}
    \For {$e \gets 1$ to $E_{cl}$}
			\State $A$ = Inference($TR_{cl}$, $\mathrm{SNN_{f}}$)
                \State ($W_{l}$, $V_{l}$) = BPTT.train ($\mathrm{SNN_{l}}$, $A$ $\cup$ $LR$, ($W_{l}$, $V_{l}$), $\eta$) 
        \EndFor
  \end{algorithmic}
          \label{alg:training}

\end{algorithm}

\subsection{Optimization of \gls{LR} memory}

To keep memory of previously learned information, an additional storage is required for the \glspl{LR}. The problem becomes increasingly serious when more classes or scenarios are learned sequentially, making the required memory grow over time. To limit the phenomenon, we introduce a time-domain lossy compression on \glspl{LR}.

\begin{figure}[t]
		\centering
		\includegraphics[width=0.8\columnwidth]{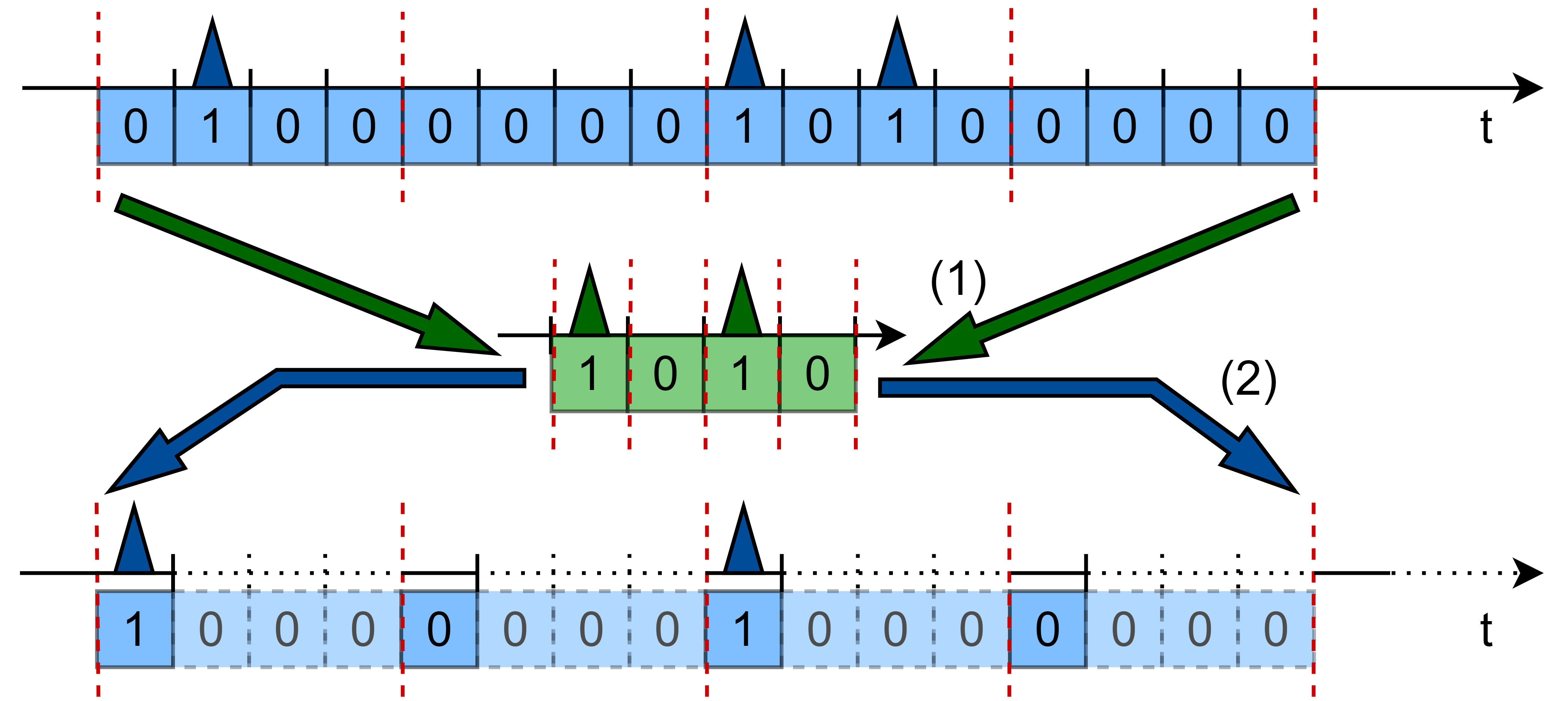}
		\caption{Example of a lossy compression (1) to store LRs. Compression ratio is here set to 4:1. Shrinking LRs and activations reduces the memory by 4$\times$. A de-compression step (2) is required to respect the SNN time scale.} 
		\label{fig:timecompression}
	\end{figure}

Fig. \ref{fig:timecompression} illustrates the compression technique applied to a generic spiking sequence. Given a compression ratio ($C_r$), each component of a \glspl{LR} sample is compressed into a sequence of $T_s / C_r$ binary activations. The algorithm involves two steps: first, the uncompressed sequence is divided into chunks composed of $C_r$ activations. Subsequently, each chunk is compressed in a single time step using a lossy compression based on thresholding: if the number of spikes within the chunk reaches a given \textit{threshold} (e.g., 1), a spike is generated in the compressed sequence. 
%
%The impact on computational resources manifests in reducing the memory required \glspl{LR} by $C_r \times$.
The impact on computational resources manifests as a $C_r \times$ memory reduction fo the \glspl{LR}.

When processing a \gls{LR} with the compressed data, decompression is needed to respect the time scale of the model. To avoid this step, we attempted to change the model's time constants, adapting them to the compressed duration of the spike sequences. However, the accuracy loss was dramatic, especially with compressed \gls{LR} performed in the first layers. We hypothesize that the inter-layer recurrent connections in these layers were tuned to a specific time scale and did not adapt well to its variation. In support of this idea, we observed that a compressed \gls{LR} on the last layer, which doesn't contain any feedback connection, almost doesn't affect the final accuracy. The solution, which also works for recurrent layers, is to uncompress the sequence by interleaving the compressed samples with zeroes to match the time scale of the spikes' sequence at run-time. 

% RE-INSERT?
%Our technique allows to preserve most of the temporal content of spike sequences, while allowing for easy decompression at run-time. Alternatively, latent spikes can be aggregated by storing them as active spike counts. While this drastically reduces the required memory to $log_2(N)$ bits/sequence, the time information is lost. Furthermore, this complicates the sequence expansion.

% ------------ EXPERIMENTAL RESULTS ------------
\section{Experimental Results}
  
    \subsection{Experimental Setup}

We evaluated our methodology using the \glspl{SHD} dataset \cite{9311226}, which comprises 10,420 spiking trains of 100 timesteps each. These trains consist of audio samples obtained through a conversion system inspired by the human cochlea. The dataset is categorized into 20 classes, corresponding to numbers 0 to 9, pronounced by 12 speakers in both English and German.
We employed a recurrent \gls{SNN} with 700 input neurons and 20 outputs (one for each class), including 4 layers with decreasing output size (200-100-50-20). As described in Section \ref{sec:snn_background}, we adopted a Synaptic Conductance-based Second Order \gls{LIF} neuron model, trained using \gls{BPTT} and a fast-sigmoid \gls{SG}. Our training setup incorporates a learning rate of $\eta=10^{-3}$ for the Sample Incremental task and $\eta=2 \cdot 10^{-4}$ for the Class Incremental task.

    \subsection{Weights initialization}

    When pre-training an \glspl{SNN} for a Class-Incremental setup, neurons associated with unlearned classes are trained to be inactive. Therefore, when adding a new class to the pre-trained classifier, the yielded accuracy is poor, reaching a maximum of 57\%. This issue can be addressed by re-initializing the neuron weights devoted to detect the new class. When performing weight re-initialization, the obtained accuracy is strictly linked to the re-initialization strategy. Notably, random, constant, and Xavier-Glorot initializations \cite{pmlr-v9-glorot10a} proved inefficient for proficient learning of the new class. Instead, we adopted a normal random distribution with mean and variance aligned with those of the classifier's weights trained on the old classes. This approach enables our model to achieve a remarkable 92.9\% accuracy on the new class. Further details are discussed in Section \ref{subsec:class_inc_res}.  Unlike the Class-Incremental scenario, the Sample-Incremental scenario does not require any initialization, as the number of classes does not change between samples.

        \begin{figure}[!ht]
            \centering

            \includegraphics[width=\columnwidth]{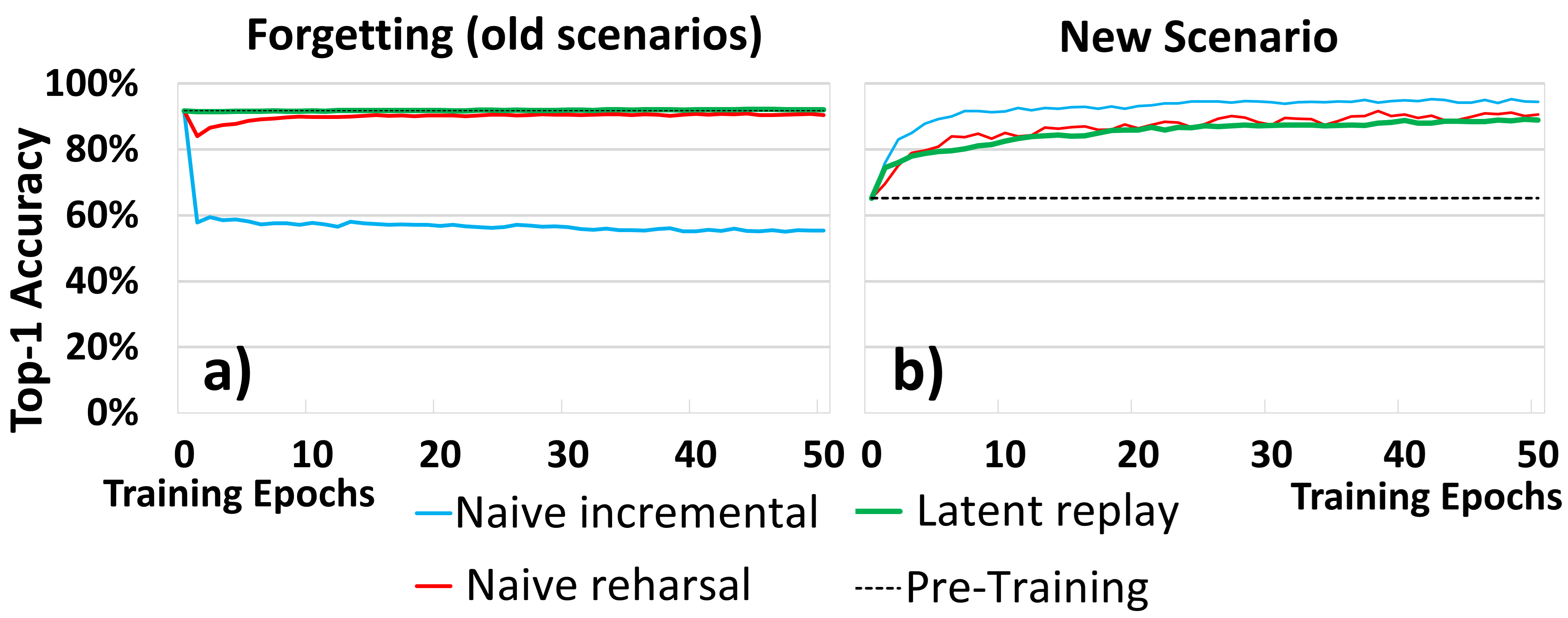}
            \caption{Measurement of (a) forgetting and (b) Top-1 accuracy on the new scenario (Sample-Incremental CL). Latent Replays are applied to the 2$^{nd}$ to last layer of our model using 2,560 past samples for the replay.}
            \label{fig:forgetting_vs_new}
        \end{figure}
        
    \subsection{Sample-Incremental CL}

Our model undergoes pre-training on 11 out of 12 scenarios (speakers) to simulate user personalization in keyword spotting tasks. The 12$^{th}$ scenario is introduced using a Sample-Incremental \gls{CL} approach. To benchmark our technique against methodologies like \cite{proietti2023memory}, we conduct experiments in three setups: (i) a \textit{na\"ive incremental} setup, where the 12$^{th}$ speaker is learned without rehearsal; (ii) a \textit{na\"ive rehearsal} setup, mixing the 12$^{th}$ speaker's input activations with 2,560 samples from the training step; (iii) the proposed \gls{LR} setup, integrating 2,560 \glspl{LR} stored at the output of the 2$^{nd}$-to-last layer. Fig. \ref{fig:forgetting_vs_new}-a) displays the forgetting measurement, represented by the Top-1 accuracy of the \gls{SNN} after 50 \gls{CL} epochs, focusing on the old scenarios. Fig. \ref{fig:forgetting_vs_new}-b) showcases the accuracy of the new scenario.

As observed in \glspl{CNN}, \glspl{SNN} face Catastrophic Forgetting when learning new samples without rehearsal. In na\"ive incremental, few epochs specialize the model toward the new speaker ($>$ 90\% accuracy), resulting in almost 40\% accuracy loss on the old speakers. In na\"ive rehearsal, rehearsal data enable the model to retain nearly the same accuracy before the \gls{CL} routine, limiting the accuracy drop on the old speakers to a 2\%. Conversely, the new speaker is learned with almost 90\% accuracy.
% Hence, rehearsal data minimize forgetting while the new speaker is learned with a Top-1 accuracy 5\% lower than na\"ive incremental, as the speaker doesn't specialize in a single scenario. 
Retraining only the last 2 layers with \glspl{LR} maximizes accuracy retention on old scenarios. The pre-training accuracy of 11 speakers is enhanced by 1\%, as the new speaker's samples strengthen classification in classes seen during pre-training. The Top-1 accuracy on the new speaker is increased by 24\%, reaching an overall $88\%$. This is 2\% lower than naïve rehearsal, but it is associated with a 7$\times$ reduction in the memory required to store rehearsal data.

 \begin{figure}[!ht]
            \centering

        \includegraphics[width=\columnwidth]{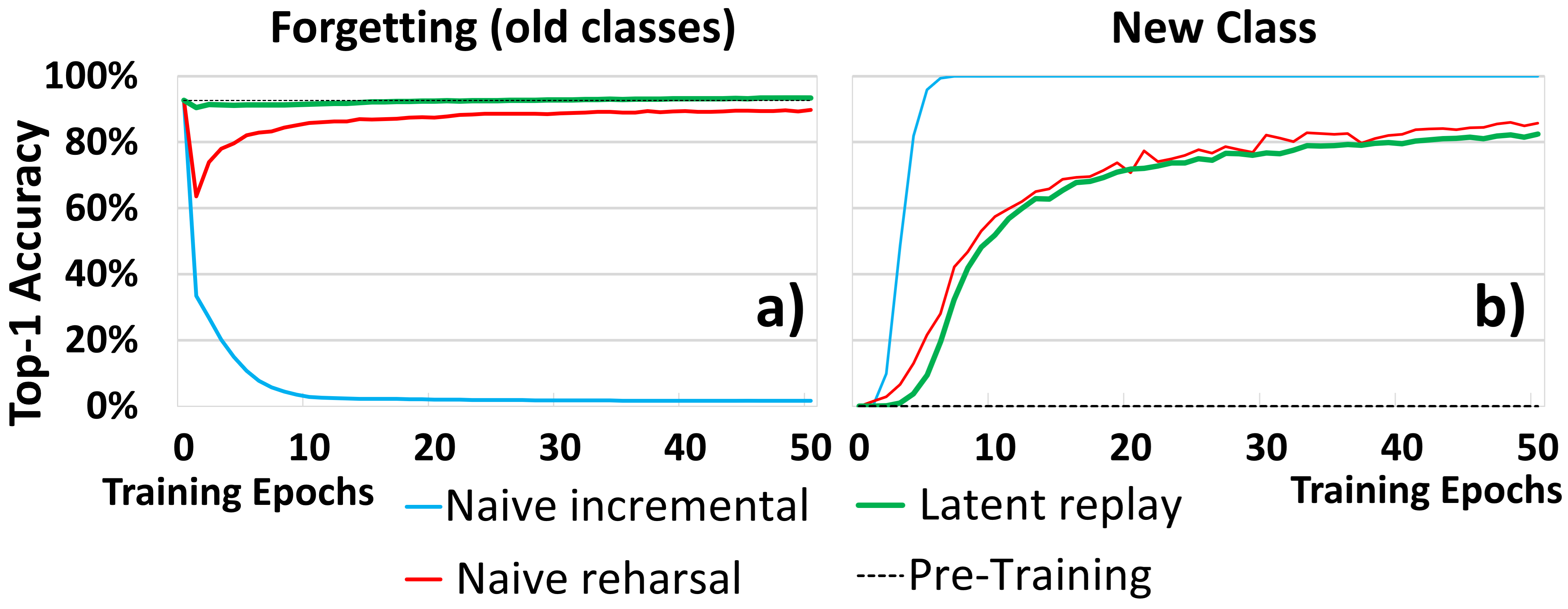}
        \caption{Measurement of (a) forgetting and (b) Top-1 accuracy on the new class (Class-Incremental CL). Latent Replays are applied to the 2$^{nd}$ to the last layer of our model. }
        \label{fig:merged_class}
        \end{figure}

\subsection{Class-Incremental CL}
    \label{subsec:class_inc_res}

In this scenario, we simulate the introduction of a new keyword by pre-training our model on 19 out of 20 classes and adding the 20$^{th}$ class subsequently. Similar to the Sample-Incremental setup, we compare (i) a \textit{na\"ive incremental}, (ii) a \textit{na\"ive rehearsal}, and (iii) \textit{the proposed \gls{LR} setup}. Fig. \ref{fig:merged_class}-a) measures forgetting, while \ref{fig:merged_class}-b) shows Top-1 accuracy on the new class. 
In terms of forgetting, \glspl{LR} show no accuracy drop, outperforming both na\"ive rehearsal ($\simeq 3\%$ drop) and na\"ive incremental (complete forgetting).  
The new class is learned within 50 epochs for all three methods. While na\"ive incremental overfits the new class, completely forgetting the past ones, \glspl{LR} achieve the highest accuracy of 83\% without notable forgetting. Again, the accuracy reached by na\"ive rehearsal is a bit higher on the new data, and exceeds \glspl{LR} by 3\%. However this comes at the cost of a 3\% forgetting of the past classes, with a $7 \times$ memory occupation.

\subsection{Classification Accuracy vs Number of LRs}

        \begin{figure}[t]
            \centering

            \includegraphics[width=\columnwidth]{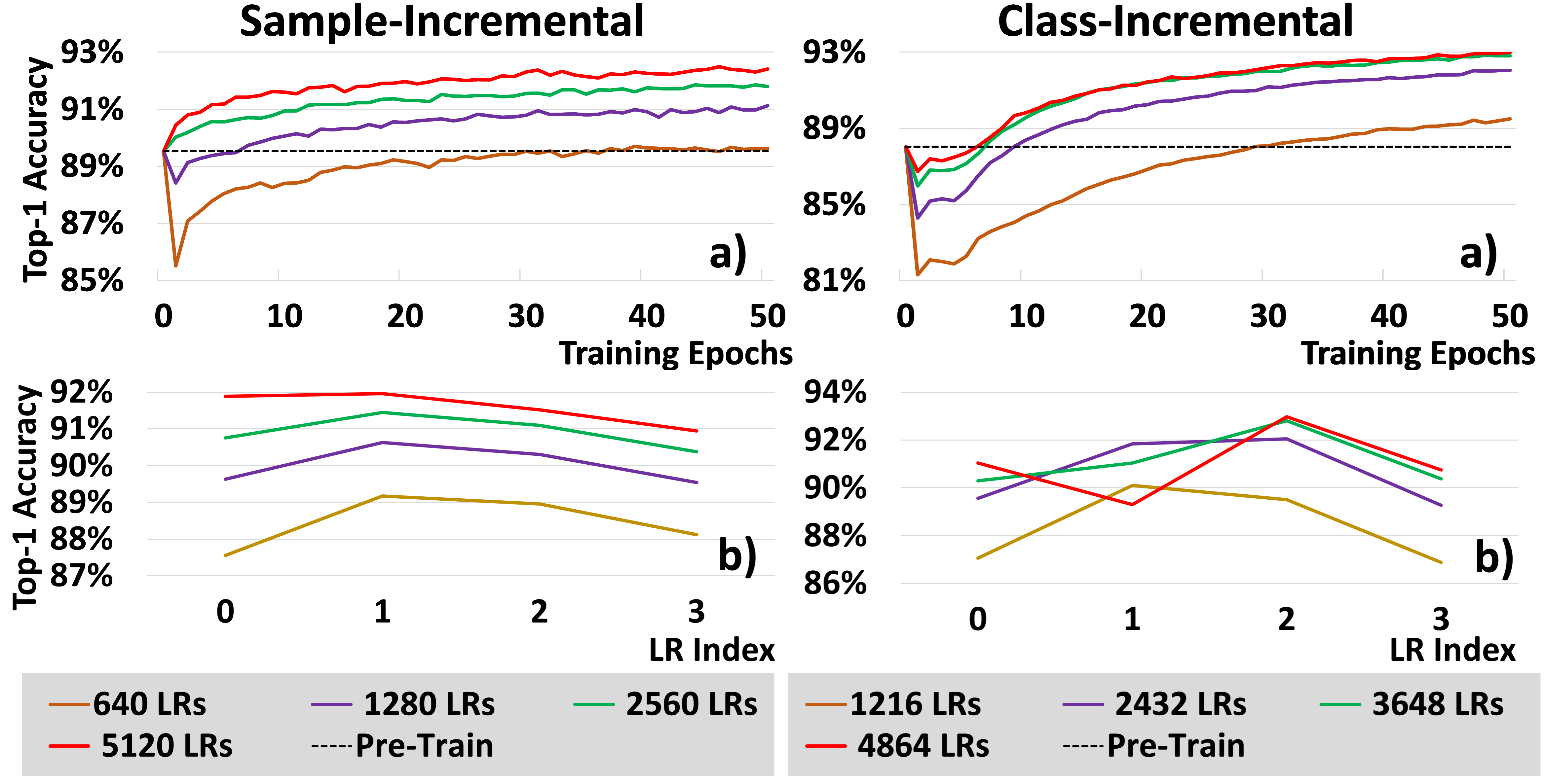}
            \caption{Top-1 accuracy on SHD's test set in Sample and Class-Incremental CL, in case of a variable number of LRs (a) with respect to the epochs, (b) with respect to the LR index.}
            \label{fig:latent_analysis}
    \end{figure}

Fig. \ref{fig:latent_analysis} comprehensively analyzes the impact of various hyperparameters on accuracy, examining (a) the number of \glspl{LR} and (b) \gls{LR} indices across different epochs.
In the Sample-Incremental setup, increasing \glspl{LR} positively influences Top-1 accuracy. The primary effect is on the maximum accuracy, rather than the convergence slope, which remains relatively constant across all cases. An initial forgetting can be observed with 640 and 1280 past samples, becoming more and more accentuated with decreasing numbers of \glspl{LR}. However, in all the cases the network is able to retrieve the past information, bringing the accuracy back to its initial value, or exceeding it. The choice of \glspl{LR} index impacts maximum accuracy as well: figure \ref{fig:latent_analysis} shows the superiority of \glspl{LR} (indexes 1, 2 and 3) with respect to na\"ive reharsal (index 0). The highest accuracy of 92.5\% is achieved with 5,120 \glspl{LR} at layer index 1, as layer 0 retains knowledge of old data while extending towards old and new samples.

Increasing \glspl{LR} has similar effects for the Class-Incremental \gls{CL} setup. While varying the layer index, we observe a peak accuracy at \glspl{LR} index 2 , making Top-1 accuracy reach 93\% with 4,864 \glspl{LR}. This indicates the capability of the final layers to learn the new class, exploiting the higher-level features learned by previous layers. An interesting effect can be noted at index 1, where the curve is non-monotonic: in this case, a larger number of \glspl{LR} delays the convergence of the model, preventing it from reaching an acceptable accuracy within the 50 epochs.

\subsection{Compressed LRs}

\begin{table}[t]
        \centering
        %\caption{Memory footprint of a single LR, with several compression ratios $C_r$, when stored at different indices.}
        \caption{Memory-Accuracy tradeoff for Sample-Incremental CL, with 2560 LRs and variable $C_r$ and LR index. Pre-training accuracy on the scenario to be learned was 65\%.}
        \begin{tabular}{|c|c|c|c|c|c|}

            \hline
            \multicolumn{2}{|c|}{\backslashbox{\textbf{$C_r$}}{\textbf{LRs}}}
            & \textbf{Na\"ive rehearsal}
            & \textbf{Layer 1}
            & \textbf{Layer 2}
            & \textbf{Layer 3}                   \\
            
            \hline
            \multirow{2}{*}{\textbf{1:1}}   
            & \textbf{Acc}
            & 90.50\%
            & \textbf{92.46\%}
            & 91.79\%
            & 90.37\%                   \\
            \cline{2-6}
            & \textbf{Mem}
            & 22.4 MB 
            & \textbf{6.4 MB}
            & 3.2 MB
            & 1.6 MB                   \\
            \hline

            %& 70000 b = 8750 B -> 22.400.000 B
            %& 20000 b = 2500 B -> 6.400.000 B
            %& 10000 b = 1250 B -> 3.200.000 B
            %& 5000  b = 625 B  -> 1.600.000 B

            \hline
            \multirow{2}{*}{\textbf{1:5}}
            & \textbf{Acc}
            & 67.81\%
            & 79.71\%
            & 89.91\%
            & \textbf{90.27\%}                   \\
            \cline{2-6}
            &\textbf{Mem}   
            & 4.48 MB
            & 1.28 MB
            & 640 KB
            & \textbf{320 KB}                      \\
            \hline     

            %& 14000 b = 1750 B -> 4.480.000 B
            %& 4000 b  = 500 B -> 1.280.000 B
            %& 2000 b  = 250 B -> 640.000 B
            %& 1000 b  = 125 B -> 320.000 B

            \hline
            \multirow{2}{*}{\textbf{1:10}}
            & \textbf{Acc}   
            & 67.12\%
            & 68.12\%
            & 84.73\%
            & \textbf{88.79\%}                   \\
            \cline{2-6}
            &\textbf{Mem}   
            & 2.24 MB
            & 640 KB
            & 320 kB
            & \textbf{160 kB}                      \\
            \hline

            %& 7000 b = 875 B -> 2.240.000 B
            %& 2000 b = 250 B -> 640.000 B 
            %& 1000 b = 125 B -> 320.000 B
            %& 500  b = 62,5 B -> 160.000 B
            
        \end{tabular}
        \label{tab:memacc_sample}
        \vspace{0.5cm}
    \end{table}

    \begin{table}[t]
        \centering
        %\caption{Memory footprint of a single LR, with several compression ratio $C_r$, when stored at different indices.}
        \caption{Memory-Accuracy tradeoff for Class-Incremental CL, with 2432 LRs and variable $C_r$ and LR index.}
        \begin{tabular}{|c|c|c|c|c|c|}

            \hline
            \multicolumn{2}{|c|}{\backslashbox{\textbf{$C_r$}}{\textbf{LRs}}}
            & \textbf{Na\"ive rehearsal}
            & \textbf{Layer 1}
            & \textbf{Layer 2}
            & \textbf{Layer 3}                   \\

            %\multicolumn{2}{|c|}{}{}
            %& \textbf{rehearsal}
            %& 
            %& 
            %&           \\
            
            \hline
            \multirow{2}{*}{\textbf{1:1}}   
            & \textbf{Acc}
            & 89.55\%
            & 91.83\%
            & \textbf{92.05\%}
            & 89.26\%                   \\
            \cline{2-6}
            & \textbf{Mem}
            & 22.4 MB 
            & 6.4 MB
            & \textbf{3.2 MB}
            & 1.6 MB                   \\
            \hline

            %& 70000 b = 8750 B -> 22.400.000 B
            %& 20000 b = 2500 B -> 6.400.000 B
            %& 10000 b = 1250 B -> 3.200.000 B
            %& 5000  b = 625 B  -> 1.600.000 B

            \hline
            \multirow{2}{*}{\textbf{1:5}}
            & \textbf{Acc}
            & 84.93\%
            & 57.43\%
            & 82.03\%
            & \textbf{86.78\%}                   \\
            \cline{2-6}
            &\textbf{Mem}   
            & 4.48 MB
            & 1.28 MB
            & 640 KB
            & \textbf{320 KB}                      \\
            \hline     

            %& 14000 b = 1750 B -> 4.480.000 B
            %& 4000 b  = 500 B -> 1.280.000 B
            %& 2000 b  = 250 B -> 640.000 B
            %& 1000 b  = 125 B -> 320.000 B

            \hline
            \multirow{2}{*}{\textbf{1:10}}
            & \textbf{Acc}   
            & 77.43\%
            & 30.75\%
            & 69.42\%
            & \textbf{85.53\%}                   \\
            \cline{2-6}
            &\textbf{Mem}   
            & 2.24 MB
            & 640 KB
            & 320 kB
            & \textbf{160 kB}                      \\
            \hline

            %& 7000 b = 875 B -> 2.240.000 B
            %& 2000 b = 250 B -> 640.000 B 
            %& 1000 b = 125 B -> 320.000 B
            %& 500  b = 62,5 B -> 160.000 B
            
        \end{tabular}
        \label{tab:memacc_class}
    \end{table}

We now explore the memory-accuracy trade-off of our \gls{LR} compression algorithm. 
The choice of the \glspl{LR}' layer index and the time compression factor $C_r$ is based on the trade-off between accuracy and memory requirements.
Tables \ref{tab:memacc_sample} and \ref{tab:memacc_class} present the results obtained on the full SHD test set, for Sample and Class-Incremental setups respectively.
In the case of \textit{na\"ive rehearsal}, the rehearsal data is stored as past input activations, consisting of 100 timesteps and 700 inputs. 
The spatial size of LRs aligns with the size of the layer on which they are stored as input, such as 100 inputs for layer 2. After selecting a $C_r$, the rehearsal data is additionally compressed on the time axis; for example, with $C_r = 5$, the data is reduced by 5$\times$, featuring 20 timesteps. Therefore, all the data is scaled proportionally. Our experiments show a minimal size of the rehearsal data, with LRs collected on layer 3 using $C_r = 10$, being 140$\times$ smaller than \textit{na\"ive rehearsal} without compression.
For the Sample-Incremental setup (Tab. \ref{tab:memacc_sample}), we observe that a Top-1 accuracy of 92.46\% is reached for uncompressed \glspl{LR} with index 1, surpassing the na\"ive rehearsal by 2\%. Compressed LRs reduce the maximum accuracy: in case of a 1:5 ratio, 90.27\% is achieved with index 3, while the accuracy drop on earlier LR indices becomes prohibitive before layer 2. This drop is accentuated for 1:10 compression, achieving a Top-1 accuracy of 88.79\% on index 3. However, a memory save of 10$\times$ is attested. 

Also in case of a Class-Incremental setup (Tab. \ref{tab:memacc_class}), the Top-1 accuracy of Latent-Replay, obtained at \gls{LR} index 2, surpasses the Naive-Incremental by a 2.5\%. Performing a \gls{LR} at index 3 brings back the accuracy to 89.26\%, comparable with Na\"ive reharsal, but with a memory requirement $14\times$ lower. Finally 1:5 and 1:10 time compressions correspond to a further drop of 2.5\% and 3.7\%, but with a memory saving of $5\times$ and $10\times$ . The accuracy loss for previous indexes is still too high, indicating that a lighter compression is required.

\color{black}
\subsection{Comparison with other compressions}

Our technique focuses on preserving most of the temporal content of spike sequences, while allowing for an easy decompression at run-time. Alternatively, latent spikes can be aggregated by storing them as active spike counts. 

%We test two additional compression methods: in the first, we aggregate spikes over the full sequence and expand during rehearsal, putting them all in the first time-steps. This causes a 12\% to 14\% accuracy drop compared to our technique, with compression ratios of 10 and 20.
We test two additional compression methods, which we compare with our sub-sampling approach (Fig.\ref{fig:timecompression}): first, we aggregate spikes over the full sequence and expand during rehearsal, placing all spikes in the first time-steps. While this drastically reduces the required memory to $log_2(N)$ bits/sequence, the time information is lost, causing a 12\% to 14\% accuracy drop compared to our technique, with $C_r$ = 10 and 20. 
%
%A different decompression approach, such as equally spaced spikes, might perform better, but requires further analysis. 
%
The second is a hybrid approach: we divide sequences into chunks and accumulate spikes within each chunk. This leads to a 2\% accuracy improvement with respect to our method, at the cost of a $3 \times$ and $4 \times$ memory occupation with $C_r = 20$ and $C_r = 10$, respectively.
%Furthermore, both the technique complicate the sequence expansion. 
%
%The choice depends on the requirements, but \glspl{SNN} seem robust to forgetting even with approximated versions of past latent activations.
Therefore, for the considered keyword spotting setup, our method provides an optimal trade-off between accuracy, memory occupation and run-time decompression complexity. Also, the temporal properties of the spike trains prove to be more relevant than the absolute number of spikes.
\color{black}

\subsection{Multi-Class-Incremental Setup}
    
    \begin{figure}[t]
            \centering

        \includegraphics[width=\columnwidth]{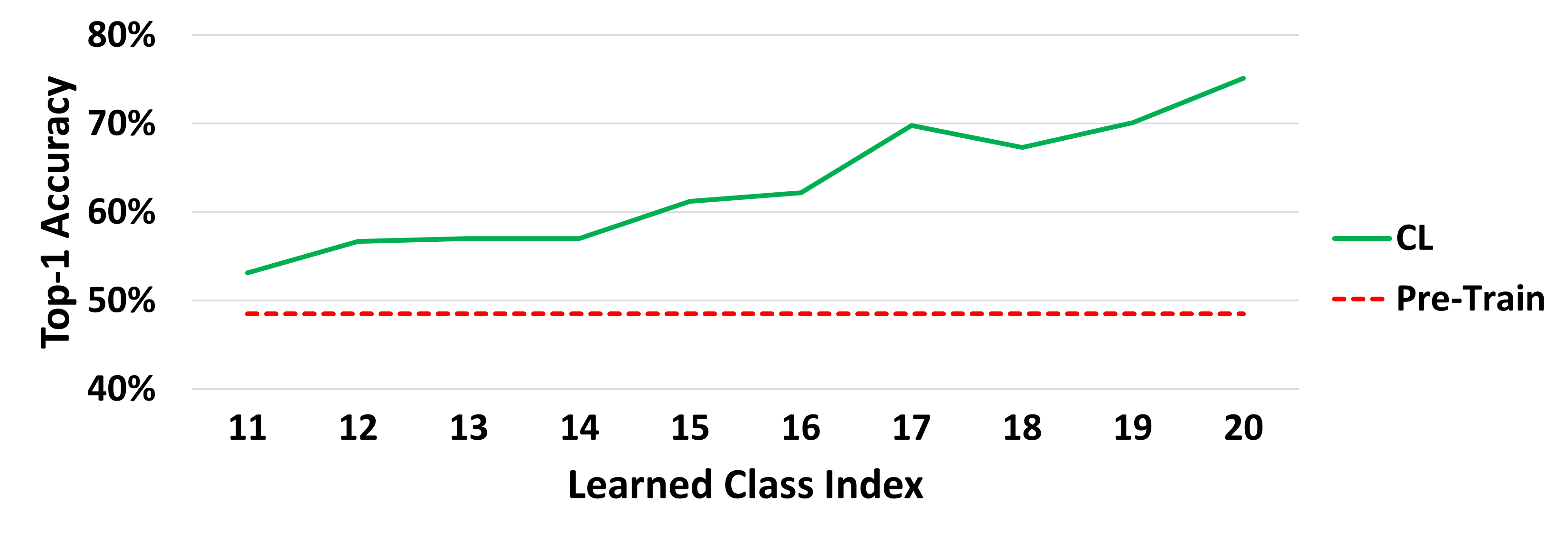}
        \caption{Multi-Class-Incremental CL on SHD. Here, a pre-trained model learns to classify 10 more classes, starting from a pre-training on 10 classes.}
        \label{fig:multiclass}
    \end{figure}

    We now provide a more complex testing scenario, i.e., a keyword spotting language shift in which our model learns to classify 10 classes of digits in German, starting from 10 pre-learned English-spoken digits. The pre-trained knowledge is stored as 2560 \glspl{LR}, 256 per-class. After each new class is learned, 256 \glspl{LR} are added to the memory. Our model is trained for 50 epochs per class, with $C_r = 2$.

    Fig. \ref{fig:multiclass} shows the Top-1 accuracy over the 20-class SHD test set. The pre-training accuracy is of 48.5\%, corresponding to an accuracy of 97\% on the first 10 classes only. While learning with CL, a monotonic growth is assessed, learning each successive class with 88.2\% average accuracy. For each new class learned, we observe an average forgetting of 2.2\% on all the previous classes. In the end, we observe a final accuracy of 78.4\% on the full test set.

% ------------ CONCLUSION ------------
%\section*{Conclusion}

%This work presented the first memory-efficient Rehearsal-based \gls{CL} solution on SNNs, inspired by the Latent Replay algorithm. Results on the keyword spotting Spiking Heidelberg Dataset showed a a top-1 accuracy of 92.5\% and of 92\%, on a Sample-Incremental and Class-Incremental, while requiring 6.4 MB and 3.2 MB, respectively.
%
%Then, taking advantage from the spiking data structure, we proposed a user-tunable lossy, temporal domain compression algorithm, which, by selecting the most performing Latent Replay index, reduced the memory requirements down to 160 kB, while limiting the accuracy loss to 4\%, with respect to the most na\"ive approach.
%
%Moreover, in a Multi-Class-Incremental setup, we achieved 78.4\% accuracy with compressed rehearsal data, while learning a set of 10 new keywords, paving the way to a new low-power, high-accuracy approach for Continual Learning on edge.

% ------------ BIBLIOGRAPHY ------------
\bibliographystyle{plain}
\bibliography{bibliography}

\begin{thebibliography}{10}

\bibitem{10098596}
Ayoub~Benali Amjoud and Mustapha Amrouch.
\newblock Object detection using deep learning, cnns and vision transformers: A review.
\newblock {\em IEEE Access}, 11:35479--35516, 2023.

\bibitem{amodei2016concrete}
Dario Amodei, Chris Olah, Jacob Steinhardt, Paul Christiano, John Schulman, and Dan Man{\'e}.
\newblock Concrete problems in ai safety.
\newblock {\em arXiv preprint arXiv:1606.06565}, 2016.

\bibitem{brette2005}
Romain Brette and Wulfram Gerstner.
\newblock Adaptive exponential integrate-and-fire model as an effective description of neuronal activity.
\newblock {\em Journal of Neurophysiology}, 94(5):3637--3642, 2005.

\bibitem{9311226}
Benjamin Cramer, Yannik Stradmann, Johannes Schemmel, and Friedemann Zenke.
\newblock The heidelberg spiking data sets for the systematic evaluation of spiking neural networks.
\newblock {\em IEEE Transactions on Neural Networks and Learning Systems}, 33(7):2744--2757, 2022.

\bibitem{ehret2021continual}
Benjamin Ehret, Christian Henning, Maria~R. Cervera, Alexander Meulemans, Johannes von Oswald, and Benjamin~F. Grewe.
\newblock Continual learning in recurrent neural networks, 2021.

\bibitem{eshraghian2023}
Jason~K. Eshraghian, Max Ward, Emre~O. Neftci, Xinxin Wang, Gregor Lenz, Girish Dwivedi, Mohammed Bennamoun, Doo~Seok Jeong, and Wei~D. Lu.
\newblock Training spiking neural networks using lessons from deep learning.
\newblock {\em Proceedings of the IEEE}, 111(9):1016--1054, 2023.

\bibitem{9731734}
Charlotte Frenkel and Giacomo Indiveri.
\newblock Reckon: A 28nm sub-mm2 task-agnostic spiking recurrent neural network processor enabling on-chip learning over second-long timescales.
\newblock In {\em 2022 IEEE International Solid-State Circuits Conference (ISSCC)}, volume~65, pages 1--3, 2022.

\bibitem{pmlr-v9-glorot10a}
Xavier Glorot and Yoshua Bengio.
\newblock Understanding the difficulty of training deep feedforward neural networks.
\newblock In Yee~Whye Teh and Mike Titterington, editors, {\em Proceedings of the Thirteenth International Conference on Artificial Intelligence and Statistics}, volume~9 of {\em Proceedings of Machine Learning Research}, pages 249--256, Chia Laguna Resort, Sardinia, Italy, 13--15 May 2010. PMLR.

\bibitem{Golden688622}
Ryan Golden, Jean~Erik Delanois, Pavel Sanda, and Maxim Bazhenov.
\newblock Sleep prevents catastrophic forgetting in spiking neural networks by forming joint synaptic weight representations.
\newblock {\em bioRxiv}, 2020.

\bibitem{hammouamri:hal-03887417}
Ilyass Hammouamri, Timoth{\'e}e Masquelier, and Dennis Wilson.
\newblock {Mitigating Catastrophic Forgetting in Spiking Neural Networks through Threshold Modulation}.
\newblock {\em {Transactions on Machine Learning Research Journal}}, November 2022.

\bibitem{10.1007/978-3-030-58607-2_23}
Bing Han and Kaushik Roy.
\newblock Deep spiking neural network: Energy efficiency through time based coding.
\newblock In Andrea Vedaldi, Horst Bischof, Thomas Brox, and Jan-Michael Frahm, editors, {\em Computer Vision -- ECCV 2020}, pages 388--404, Cham, 2020. Springer International Publishing.

\bibitem{han2023adaptive}
Bing Han, Feifei Zhao, Wenxuan Pan, Zhaoya Zhao, Xianqi Li, Qingqun Kong, and Yi~Zeng.
\newblock Adaptive reorganization of neural pathways for continual learning with spiking neural networks, 2023.

\bibitem{Kemker_McClure_Abitino_Hayes_Kanan_2018}
Ronald Kemker, Marc McClure, Angelina Abitino, Tyler Hayes, and Christopher Kanan.
\newblock Measuring catastrophic forgetting in neural networks.
\newblock {\em Proceedings of the AAAI Conference on Artificial Intelligence}, 32(1), Apr. 2018.

\bibitem{liang2021pruning}
Tailin Liang, John Glossner, Lei Wang, Shaobo Shi, and Xiaotong Zhang.
\newblock Pruning and quantization for deep neural network acceleration: A survey.
\newblock {\em Neurocomputing}, 461:370--403, 2021.

\bibitem{lomonaco2020rehearsal}
Vincenzo Lomonaco, Davide Maltoni, Lorenzo Pellegrini, et~al.
\newblock Rehearsal-free continual learning over small non-iid batches.
\newblock In {\em CVPR Workshops}, volume~1, page~3, 2020.

\bibitem{neftci2019}
Emre~O. Neftci, Hesham Mostafa, and Friedemann Zenke.
\newblock Surrogate gradient learning in spiking neural networks: Bringing the power of gradient-based optimization to spiking neural networks.
\newblock {\em IEEE Signal Processing Magazine}, 36(6):51--63, 2019.

\bibitem{9341460}
Lorenzo Pellegrini, Gabriele Graffieti, Vincenzo Lomonaco, and Davide Maltoni.
\newblock Latent replay for real-time continual learning.
\newblock In {\em 2020 IEEE/RSJ International Conference on Intelligent Robots and Systems (IROS)}, pages 10203--10209, 2020.

\bibitem{proietti2023memory}
Michela Proietti, Alessio Ragno, Roberto Capobianco, et~al.
\newblock Memory replay for continual learning with spiking neural networks.
\newblock In {\em 2023 IEEE 33rd International Workshop on Machine Learning for Signal Processing (MLSP)}, pages 1--6, 2023.

\bibitem{9586281}
Rachmad Vidya~Wicaksana Putra and Muhammad Shafique.
\newblock Spikedyn: A framework for energy-efficient spiking neural networks with continual and unsupervised learning capabilities in dynamic environments.
\newblock In {\em 2021 58th ACM/IEEE Design Automation Conference (DAC)}, pages 1057--1062, 2021.

\bibitem{Rebuffi_2017_CVPR}
Sylvestre-Alvise Rebuffi, Alexander Kolesnikov, Georg Sperl, and Christoph~H. Lampert.
\newblock icarl: Incremental classifier and representation learning.
\newblock In {\em Proceedings of the IEEE Conference on Computer Vision and Pattern Recognition (CVPR)}, July 2017.

\bibitem{10.3389/fncom.2022.1037976}
Nicolas Skatchkovsky, Hyeryung Jang, and Osvaldo Simeone.
\newblock Bayesian continual learning via spiking neural networks.
\newblock {\em Frontiers in Computational Neuroscience}, 16, 2022.

\bibitem{werbos1990}
P.J. Werbos.
\newblock Backpropagation through time: what it does and how to do it.
\newblock {\em Proceedings of the IEEE}, 78(10):1550--1560, 1990.

\bibitem{Yin2021}
Bojian Yin, Federico Corradi, and Sander~M. Bohte.
\newblock Accurate and efficient time-domain classification with adaptive spiking recurrent neural networks, Mar 12 2021.

\end{thebibliography}

\end{document}